\documentclass[conference]{IEEEtran}
\IEEEoverridecommandlockouts
\usepackage{cite}
\usepackage{amsmath,amssymb,amsfonts}
\usepackage{algorithm}
\usepackage[noend]{algorithmic}
\usepackage{graphicx}
\usepackage{textcomp}
\usepackage{multirow}
\usepackage{xcolor}
\def\BibTeX{{\rm B\kern-.05em{\sc i\kern-.025em b}\kern-.08em
    T\kern-.1667em\lower.7ex\hbox{E}\kern-.125emX}}

\newtheorem{definition}{Definition}

\begin{document}

\title{From domain-landmark graph learning to problem-landmark graph generation\\
}

 \author{
 \IEEEauthorblockN{1\textsuperscript{st} Cristian Pérez-Corral}
 \IEEEauthorblockA{\textit{Universitat Politècnica de València} \\
 Valencia, Spain \\
 cpercor@upv.es}
 \and
 \IEEEauthorblockN{2\textsuperscript{nd} Antonio Garrido}
 \IEEEauthorblockA{\textit{VRAIN} \\
 \textit{Universitat Politècnica de València}\\
 Valencia, Spain \\
 agarridot@dsic.upv.es}
 \and
 \IEEEauthorblockN{3\textsuperscript{rd} Laura Sebastia}
 \IEEEauthorblockA{\textit{VRAIN} \\
 \textit{Universitat Politècnica de València}\\
 Valencia, Spain \\
 lstarin@dsic.upv.es}
 }

\maketitle

\begin{abstract}
Landmarks have long played a pivotal role in automated planning, serving as crucial elements for improving the planning algorithms.
The main limitation of classical landmark extraction methods is their sensitivity to specific planning tasks. This results in landmarks fully tailored to individual instances, thereby limiting their applicability 
across other instances of the same planning domain.
We propose a novel approach that learns 
landmark relationships from multiple planning tasks of a planning domain.
This leads to the creation of 
a \textit{probabilistic lifted ordering graph}, as a structure that captures weighted abstractions of relationships between parameterized landmarks.
Although these orderings are not 100\% true (they are probabilistic), they can still be very useful in planning.
Next, given a new planning task for that domain, we
instantiate the 
relationships from that graph to this particular instance.
This instantiation operates in two 
phases. First, it generates two graphs: the former instantiating information from the initial state and the latter from the goal state.
Second, it combines these two graphs into one unified graph by searching equivalences to extract landmark orderings.
We evaluate the 
precision and recall
of the 
information found by our approach over 
well-known planning domains.
\end{abstract}


\section{Introduction}

A landmark \cite{hoffmann2004ordered} is defined as a predicate that must appear in every plan that solves a particular planning task,
while an ordering between two landmarks defines a predecessor relationship that must always hold.
Landmarks have been  used in automated planning, mainly for search-guiding purposes \cite{richter2010lama,richter2008landmarks}, goal and plan recognition 
\cite{Pereira2020}, generalized planning \cite{segovia2022scaling}, numeric planning \cite{scala2017landmarks} and temporal planning \cite{karpas2015temporal}.
Landmarks are totally dependent on the particular planning task, i.e., different 
planning tasks of the same domain will likely have different landmarks depending on the objects involved, initial state, and goals. This entails a limitation: landmarks need to be generated from scratch for every planning task.

Although most state-of-the-art planners reason on grounded (instantiated) information, planning models  
are usually defined in a lifted way, such as PDDL, the standard Planning Domain Definition Language.
Lifted planning has recently attracted more attention. On the one hand, there is a focus on learning domain-independent properties \cite{Chen_Thiébaux_Trevizan_2024,Pereira2020}. On the other hand, there is 
a focus on extending classical heuristics 
to be used in lifted planning 
\cite{Corrêa_Pommerening_Helmert_Francès_2022,wichlacz2022landmark}.

From the landmark perspective, specific abstractions can be made in a planning domain.
For example, in the BlocksWorld domain, if the goal is to achieve \texttt{on(a,b)}, the
  landmarks (i.e., necessary conditions) would include \texttt{holding(a)} and \texttt{clear(b)}. We can generalize this 
  by using variables instead of objects: 
  \texttt{on(?x0,?x1)} needs to be preceded by the conditions \texttt{holding(?x0)} and \texttt{clear(?x1)}, where \texttt{?x0} and \texttt{?x1} are variables over the blocks defined by the planning task.  This ordering holds  in every situation: 
  in order to have one block stacked on top of another, we need to hold the former and have the latter clear. 
  Other situations entail different potential orderings.
  For example, \texttt{holding(a)} can be preceded either by \texttt{ontable(a)} or \texttt{on(a,b)}. Therefore, one pseudo landmark to achieve \texttt{holding(?x0)} would be \texttt{ontable(?x0)} in some occasions, 
  but \texttt{on(?x0,?x1)} in others (with different probabilities), depending on the state of the plan. 
  Although these probabilistic ordering relationships are not 100\% true, as they do not necessarily appear in every possible plan, 
  they: 1) are consistent in many situations of the entire domain, and 2) can be 
reused in multiple instances. 
This work focuses on learning abstract properties in the form of probabilistic relationships. 

Landmark ordering relationships can be directly 
learned 
from the planning domain. However, in many real-world scenarios the domain is not fully available. In absence of domain knowledge, we can use the result of planning tasks. 
Learning common patterns (i.e., probabilistic ordering relationships and their lifted landmarks)
from these instances
can lead to a better understanding of the preconditions necessary for achieving specific goals. 
Extracting probabilistic ordering relationships between landmarks from 
multiple instances allows us to learn generalized relationships and, somehow, extend the orderings that might happen across the same domain.

Our contributions are twofold. First, we formalize the idea of learning (and lifting) the ordering landmark relationships in a domain from a dataset of planning tasks. 
We only focus on \textit{greedy necessary orders} \cite{hoffmann2004ordered}, but extracting other orders is analogous.
We generalize each of the Landmark Generation Graph (LGG) \cite{hoffmann2004ordered} created per planning task and build a probabilistic Lifted Ordering Graph (p-LOG), which captures the information learned from the dataset. 
Second, we propose a method to generate the landmark orderings for a new planning task. 
We instantiate a probabilistic Landmark Generation Graph (p-LGG) for the new instance. The p-LGG extends the LGG with information on probabilistic 
orderings, by combining the information that can be instantiated from the (fully grounded) initial and goal states of the instance.
Loosely speaking, we move from probabilistic domain-landmark learning to probabilistic problem-landmark extraction.

\section{Planning background and terminology}
\label{sec:planning_back}

A grounded planning task over a given domain $\mathcal{D}$ is defined by the tuple 
$
\Pi^G = \langle \mathcal{F}, \mathcal{A}, \mathcal{I}, \mathcal{G}\rangle
$,
where $\mathcal{F}$ is a set of \textit{grounded} predicates, $\mathcal{A}$ is a set of actions, $\mathcal{I} \subseteq \mathcal{F}$ is a set of predicates that are true in the initial state, and $\mathcal{G} \subseteq \mathcal{F}$ is the set of predicates representing the goal. An action $a \in \mathcal{A}$ is a tuple $\langle pre(a), add(a), del(a) \rangle$, where $pre(a)$ represents the preconditions of $a$, i.e, the set of predicates that must exist in the current state for $a$ to be executed, and $add(a)$ and $del(a)$ represents the positive and negative effects of $a$, respectively. 

A lifted planning task over a given domain $\mathcal{D}$ is defined by the tuple  $\Pi^L = \langle \mathcal{P}, \mathcal{O}, \mathcal{A}, \mathcal{I}, \mathcal{G}\rangle
$, 
where $\mathcal{P}$ is a set of first-order predicates, $\mathcal{A}$ is a set of action schemas, $\mathcal{O}$ is a set of objects, $\mathcal{I}$ is the set of grounded predicates that are true in the initial state, and $\mathcal{G}$ is the set of grounded predicates that form the goal. Each predicate $p(x_1,\dots, x_n) \in \mathcal{P}$ has a list of $n \geq 0$ parameters, where each parameter represents a variable that can take different values from $\mathcal{O}$. We define $\mathcal{V}$ as the set of all variables used in $\Pi^L$. An action schema $A \in \mathcal{A}$ is a tuple $\langle params(A), pre(A), add(A), del(A) \rangle$ with a set of parameter variables $params(A)$, as well as \textit{precondition}, \textit{add} and \textit{del} effects, all of three are sets of predicates parametrized with variables from $params(A)$  \cite{wichlacz2022landmark}. 

\begin{definition}[Landmark]
\label{def:landmark}
    Given a planning task $\Pi^G = \langle \mathcal{F}, \mathcal{A}, \mathcal{I}, \mathcal{G}\rangle$, a predicate $p(o_1,\dots, o_m) \in F$ is a landmark of $\Pi^G$ if $p(o_1,\dots, o_m) \in \mathcal{I} \cup \mathcal{G}$ or for every plan $\pi = \langle a_1,\dots, a_i,\dots, a_n \rangle$ that solves $\Pi^G$ there is an $i$ such $p(o_1,\dots, o_m) \in add(a_i)$. 
\end{definition}

\begin{definition}[Lifted landmark]
    A partially grounded predicate $p(x_1,\dots, x_i,\dots, x_m)$ in $\Pi^L$, such that $x_i \in \mathcal{V} \cup \mathcal{O}$, is a lifted landmark if there exists a landmark $p(o_1,\dots, o_i,\dots, o_m)$ in $\Pi^G$ that can be instantiated from $p(x_1,\dots, x_i,\dots, x_m)$.
\end{definition}
In order to distinguish between the objects and variables of a landmark, we use the notation
$objects(p(x_1,\dots, x_i,\dots, x_m)) = \{x_i | x_i \in \mathcal{O}\}$, $variables(p(x_1,\dots, x_i,\dots, x_m)) = \{x_i | x_i \in \mathcal{V}\}$ and $params(p(x_1,\dots, x_i,\dots, x_m)) = objects(p(x_1,\dots, x_i,\dots, x_m)) \cup variables(p(x_1,\dots, x_i,\dots, x_m))$.
All the parameters in a landmark are objects. However, a landmark $p(x_1,\dots, x_i,\dots, x_m)$ is lifted when at least one of its parameters is a variable, that is, $|variables(p(x_1,\dots, x_i,\dots, x_m))| > 0$. 
Note that a disjunctive landmark, as defined in \cite{Helmert_Domshlak_2009,hoffmann2004ordered}, is a particular case of a lifted landmark. For example, the disjunctive landmark given by 
$\{\texttt{on(a,b),on(a,c)}\}$ is a particular case of the lifted landmark \texttt{on(a,?x0)}, where $\texttt{?x0}\in\{\texttt{b,c}\}$.

An ordering, or predecessor relationship, between two landmarks $L_1 \rightarrow L_2$ is 
represented as an edge $(L_1,L_2)$ in a graph $G=(V,E)$, where
$V$ is the set of vertices and $E$ is a multiset of edges.


\begin{definition}[Landmark Generation Graph]
\label{def:landmarkgg}
Given a task $\Pi^G$, a set of landmarks $\mathcal{L}$ of $\Pi^G$ and a set of orderings $O=\{(L_i, L_j) : L_i \rightarrow L_j\}$ between these landmarks, we define the Landmark Generation Graph of $\Pi^G$ as the graph $LGG = (V, E)$, where $V = \mathcal{L}$,  $E = O$.
\end{definition}

\section{Domain-landmark learning}

This section provides a way to abstract the landmarks and their orderings from a dataset of planning tasks to learn domain-dependent landmark orderings. The idea is to combine the landmark information from individual planning tasks of a given domain and learn the landmark orderings that can be reused for any other planning task of that domain. This involves the generation and combination of multiple graphs in a three-step approach.

\subsection{First Step. Generating the Lifted Ordering Graph}

Let us consider a collection of planning tasks $\{\Pi_1^G, \dots, \Pi_i^G,\dots, \Pi_n^G\}$ for a domain $\mathcal{D}$. For every $\Pi_i^G$ we calculate its LGG = $(V,E)$, with the objective of learning the ordering relations between landmarks. For every landmark $L \in V$, we automatically calculate its lifted version $L'$, where each parameter is a variable. Then, we create the
\textbf{Lifted Ordering Graph} of $L'$, as the graph LOG$_{L'} = (V_{L'}, E_{L'})$:
\begin{itemize}
        \item $V_{L'} = \{L'\} \cup \{L'_i | (L_i, L) \in E\}$
        \item $E_{L'} = \{(L'_i, L') | (L_i, L) \in E\}$
\end{itemize}
where $L'_i$ is the lifted version, automatically calculated, of $L_i$. Note that LOG$_{L'}$ represents a lifted version of the LGG for 
$L$.
By combining this graph for all the landmarks in $\Pi_i^G$, we generate the \textbf{Lifted Ordering Graph} for $\Pi_i^G$, given by $(V_{\Pi_i^G},E_{\Pi_i^G})$, where 
\begin{itemize}
        \item $V_{\Pi_i^G} = \displaystyle \bigcup V_{L'_j} | L_j' \text{ is a lifted landmark of } {\Pi_i^G}$
        \item $E_{\Pi_i^G} = \displaystyle \bigcup E_{L'_j} | L_j' \text{ is a lifted landmark of } {\Pi_i^G}$
\end{itemize}

Note that this combined graph represents a lifted version of all the landmarks in $\Pi_i^G$, which learns all the relationships in a parametrized way. 
It is important to highlight that the edges have been defined as a multiset because a parametrized ordering between the lifted landmarks $(L'_1, L'_2)$ can appear multiple times in $E_{\Pi_i^G}$. 
Therefore, each edge $e \in E_{\Pi_i^G}$ represents a tuple $\langle (L'_1, L'_2), n \rangle$. The first element
represents the ordering, and the second one is the number of times this ordering appears in the graph.

\subsection{Second Step. Generating the Weighted Lifted Ordering Graph}

This step takes all the LOGs (one per planning task $\Pi_i^G$) of the previous step and combines them in a \textbf{weighted Lifted Ordering Graph}. The idea is to merge the lifted information from the entire dataset in one single graph, where we have the ordering relationships (edges) and also the number of times each relationship appears (weight).
Hence, the weighted Lifted Ordering Graph,  w-LOG $= (V_\mathcal{D}, E_\mathcal{D})$ is given by:
 \begin{itemize}
        \item $V_{\mathcal{D}} = \displaystyle \bigcup_{\forall i} V_{\Pi^G_i}$
        \item $E_{\mathcal{D}} = \displaystyle \bigcup_{\forall i} E_{\Pi^G_i}$
\end{itemize}

This provides a way to learn the landmark ordering relations for a domain $\mathcal{D}$. Given two lifted landmarks $L'_1$ and $L'_2$, the edge $e_i$ that represents the ordering $L'_1 \rightarrow L'_2$ is labelled with its weight$_i$ for all the LOGs processed for $\mathcal{D}$.

\subsection{Third Step. Generating the Probabilistic Lifted Ordering Graph}

Since we are interested in the probability of an edge over the given dataset, we define the function $nGraph(L')$ that takes a lifted landmark $L'$ and returns the number of lifted ordering graphs of $L'$ contained in the w-LOG. Thus, the probability of an edge $e = (L_1', L_2')$ is calculated as the proportion it appears in the graphs LOG$_{L_2'}$. More particularly: 
\[\mu(e = \langle (L_1', L_2'), n \rangle) =  \frac{n}{nGraph(L_2')} \]


This allows us to generate the \textbf{probabilistic Lifted Ordering Graph}, p-LOG $= \langle V_\mathcal{D}, E_\mathcal{D}, \mu(E_\mathcal{D})\rangle$, where:
\[\mu(E_\mathcal{D}) = \{\mu(e) | e \in E_\mathcal{D}\}\]

\begin{figure}[t]
    \centering
    \includegraphics[width=1\linewidth]{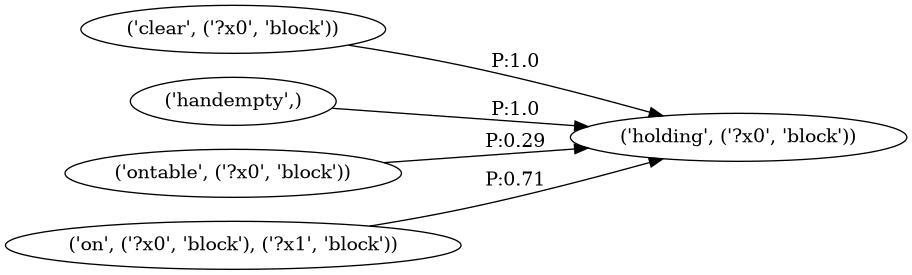}
    \caption{p-LOG for \texttt{holding(?x0)} in the BlocksWorld domain. }
    \label{fig:example-subgraph}
\end{figure}

The p-LOG represents the ordering relationships between the lifted landmarks and their probability based on the proportions in which they appear in the dataset.
Fig. \ref{fig:example-subgraph} shows an example for \texttt{holding(?x0)} in the BlocksWorld domain. For example, a probability of 0.29 in the ordering \texttt{ontable(?x0)} $\rightarrow$ \texttt{holding(?x0)} means that it appears in almost a third of the dataset. A probability of 1.0 in the ordering \texttt{clear(?x0)} $\rightarrow$ \texttt{holding(?x0)} means that it appears in the entire dataset.
It is important to note that a true ordering relationship always has probability=1.0,
provided no true orderings are missing in the dataset.
However, an ordering relationship with probability=1.0 does not necessarily imply a true ordering, as it might hold in the current dataset but not in others.

\section{Probabilistic problem-landmark graph generation}


 Let us consider a new planning task $\Pi^G$. The classical way to extract its landmarks requires applying the method proposed in \cite{hoffmann2004ordered}, which combines the information from the goal and initial states. Unfortunately, this method does not consider \textit{reusable} domain-landmark information, which would enrich the landmark extraction. On the contrary, 
 our approach uses the p-LOG calculated in the previous section, which can be 
 reused across different planning tasks of that domain. The underlying idea is to instantiate the domain-landmark information for this particular $\Pi^G$. 
 This instantiation is a two-step approach. In the first step, we generate two probabilistic LGGs (p-LGG$_\mathcal{G}$ and p-LGG$_\mathcal{I}$, starting from $\mathcal{G}$ and $\mathcal{I}$, respectively).
 In the second step, we combine these two graphs in a unique graph to finally extract the landmark orderings.

\subsection{First Step. Generating the Probabilistic Landmark Generation Graphs} 

Let us consider a p-LOG and a new planning task $\Pi^G$. The p-LGG for $\Pi^G$ needs to be instantiated from the grounded information in $\Pi^G$. 
More specifically, 
we generate two graphs p-LGG$_\mathcal{G}$ and p-LGG$_\mathcal{I}$, starting from $\mathcal{G}$ and $\mathcal{I}$, respectively.  
Intuitively, we need to create and instantiate variables in those graphs that represent the parameters of the lifted landmarks (nodes) of p-LOG.
The domain of each variable to be instantiated is formed by the objects of the variable type in $\Pi^G$. For example, if there are three blocks \texttt{a,b,c} in $\Pi^G$, the domain of a variable \texttt{?x0 - block} is initially \{\texttt{a,b,c}\}. 

Algorithm \ref{alg:generation} describes the instantiation process to generate p-LGG$_\mathcal{G}$. 
Lines 1-3 initialize the data structures. p-LGG$_\mathcal{G}$ is a map for the lifted landmarks to be instantiated and their predecessors.
For every variable $x_i$, $\mathcal{V}_{csts}[x_i]$ is a double map that stores the distinct constraints over values (objects) and variables. 
For each $x_i \in \mathcal{V}$, $\mathcal{V}_{csts}[x_i].objects$ contains the objects that $x_i$ cannot take (and need to be removed from its domain). 
$\mathcal{V}_{csts}[x_i].variables$ contains the variables that $x_i$ cannot equal.
$q$ is a queue, initialized with $\mathcal{G}$, which will process the landmarks. Lines 5-7 insert the landmark extracted from $q$ in p-LGG$_\mathcal{G}$ (if applicable). 
If $lm$ is completely lifted, i.e., all its parameters are open, its expansion will not add any new valid information and it will be  discarded. Otherwise, its predecessors are processed (lines 9-13). Line 10 updates the distinct constraints of every $x_i$ in $\mathcal{V}_{csts}[x_i]$.
For every $x_i$ in $pred$, $x_i$ must be different from the objects and variables in $lm$, provided $x_i$ is not in $lm$.
For example, let us consider the ordering $pred = (a,?x2) \rightarrow lm = (a,?x0,?x1)$.
This results in $\mathcal{V}_{csts}[?x2].objects = \{a\}$ and $\mathcal{V}_{csts}[?x2].variables = \{?x0, ?x1\}$, as $?x2$ must be distinct from $a$, $?x0$ and $?x1$.
Line 11 updates the p-LGG$_\mathcal{G}$. Lines 12-13 update the queue $q$, but only if $pred$ is not completely lifted. 


\begin{algorithm}[t]
\begin{algorithmic}[1]

\STATE p-LGG$_\mathcal{G} \leftarrow map()$
\STATE $\mathcal{V}_{csts} \leftarrow map()$
\STATE $q \leftarrow queue(\mathcal{G})$

\WHILE{$\neg q.empty$}
\STATE $lm \leftarrow q.dequeue()$
\IF{$lm \not \in \text{p-LGG}_\mathcal{G}$}
\STATE p-LGG$_\mathcal{G}[lm] \leftarrow \emptyset $
\ENDIF
\IF{$|objects(lm)| > 0 \land lm \not \in \mathcal{I}$}
\FOR{$pred \in \text{p-LOG}[lm]$}
\STATE $update\_distinct\_consts(\mathcal{V}_{csts},pred,lm)$


\STATE p-LGG$_\mathcal{G}[lm].insert(pred)$

\IF{$|objects(pred)| > 0$}
\STATE $q.enqueue(pred)$
\ENDIF
\ENDFOR
\ENDIF
\ENDWHILE
\RETURN p-LGG$_\mathcal{G}, \mathcal{V}_{csts}$
\end{algorithmic}
\caption{p-LGG$_\mathcal{G}$ generation. \textbf{input}: p-LOG, $\mathcal{G}, \mathcal{I}$. \textbf{output}: p-LGG$_\mathcal{G}$, $\mathcal{V}_{csts}$}
\label{alg:generation}
\end{algorithm}

Algorithm \ref{alg:generation} uses a backward process, starting from $\mathcal{G}$, to generate p-LGG$_\mathcal{G}$. This means that for a given ordering relation $L_1 \rightarrow L_2$, the instantiation information of the variables flows from $L_2$ to its predecessor $L_1$.
The instantiation process to generate p-LGG$_\mathcal{I}$ is analogous to this algorithm, but using a forward process that starts from $\mathcal{I}$. The information of the variables flows now from $L_1$ to its successor $L_2$. For space limitations, we do not repeat the algorithm to calculate p-LGG$_\mathcal{I}$.
The only differences are in line 3 (now $q \leftarrow queue(\mathcal{I})$), line 8 (now if $|objects(lm)| > 0 \land lm \not \in \mathcal{G}$) and lines 9-13 (the variable $pred$ is renamed as the successor $succ$).

\subsection{Second Step. Combining the Probabilistic Landmark Generation Graphs} 
\label{sec:propagation}

This step combines p-LGG$_\mathcal{G}$ and p-LGG$_\mathcal{I}$ to generate a unique p-LGG. Algorithm \ref{alg:iteprop} describes this generation.
Lines 1-6 initialize the structures: the two generated p-LGGs ($G_\mathcal{I}$ and $G_\mathcal{G}$), the predicates from $\mathcal{I}$ and $\mathcal{G}$ ($Lms_{\mathcal{I}}$ and $Lms_{\mathcal{G}}$), and the landmarks of iteration $t$ and $t+1$ ($Lms_t$ and $Lms_{t+1}$), respectively. 
The algorithm repeats while new landmarks can be extracted (line 7). Every iteration (lines 8-13) enriches the information of $G_\mathcal{I}$ with the landmarks in $Lms_{\mathcal{G}}$, and the information of $G_\mathcal{G}$ with the landmarks in $Lms_{\mathcal{I}}$. The most complex task is the instantiation of $G_\mathcal{I}$ and $G_\mathcal{G}$ (lines 9 and 11, respectively), which is described in Algorithm \ref{alg:instantiation}.
Lines 10 and 12 get the landmarks that are fully instantiated in every graph, and update $Lms_{\mathcal{I}}$ and $Lms_{\mathcal{G}}$. Finally, line 13 updates $Lms_{t+1}$.

\begin{algorithm}[t]
	\begin{algorithmic}[1]
		\STATE $G_\mathcal{I} \leftarrow \text{p-LGG}_\mathcal{I}$
		\STATE $G_\mathcal{G} \leftarrow \text{p-LGG}_\mathcal{G}$
		
		\STATE $Lms_{\mathcal{I}} \leftarrow \mathcal{I}$		
		\STATE $Lms_{\mathcal{G}} \leftarrow \mathcal{G}$		
		
		\STATE $Lms_t \leftarrow \mathcal{I} \cup \mathcal{G}$
		\STATE $Lms_{t+1} \leftarrow \emptyset$
		
		\WHILE{$Lms_{t + 1} \ne Lms_t$}
		
		\STATE $Lms_{t + 1} \leftarrow Lms_{t}$
		\STATE $G_\mathcal{I}\leftarrow instantiation(G_\mathcal{I}, Lms_{\mathcal{G}}, \mathcal{V}_{csts})$
		\STATE $Lms_{\mathcal{I}}.insert(get\_instantiated\_lms(G_\mathcal{I}))$
		\STATE $G_\mathcal{G} \leftarrow instantiation(G_\mathcal{G}, Lms_{\mathcal{I}}, \mathcal{V}_{csts})$
		\STATE $Lms_{\mathcal{G}}.insert(get\_instantiated\_lms(G_\mathcal{G}))$
		
		\STATE $Lms_{t + 1}.insert(Lms_{\mathcal{I}} \cup Lms_{\mathcal{G}})$
		\ENDWHILE
		
		\RETURN $G_\mathcal{I} \cup G_\mathcal{G}$
	\end{algorithmic}
	\caption{p-LGG combination. \textbf{input}: p-LGG$_\mathcal{I}$,p-LGG$_\mathcal{G}, \mathcal{I}, \mathcal{G}, \mathcal{V}_{csts}$. \textbf{output}: p-LGG}
	\label{alg:iteprop}
\end{algorithm}

\begin{algorithm}[t]
	\begin{algorithmic}[1]
		\STATE $var\_inst \leftarrow map()$
		\FOR{$lm \in Lms$}
		\STATE $var\_inst.update(search\_best\_equiv(\text{p-LGG},lm,\mathcal{V}_{csts}))$
		\ENDFOR
		\STATE $q \leftarrow queue(get\_lifted\_landmarks(\text{p-LGG}))$
		\WHILE{$\neg q.empty$}
		\STATE $lmLifted \leftarrow q.dequeue()$
		\STATE $lmInst \leftarrow instantiate(lmLifted, var\_inst)$
		\IF{$lmLifted \ne lmInst$}
        \IF{$lmInst \not \in \text{p-LGG}$}
        \STATE p-LGG$[lmInst] \leftarrow \emptyset$
        \ENDIF
		\STATE p-LGG$[lmInst].insert(\text{p-LGG}[lmLifted])$
		\ENDIF
		\ENDWHILE
		\RETURN p-LGG
	\end{algorithmic}\caption{Instantiation. \textbf{input} p-LGG, $Lms$, $\mathcal{V}_{csts}$. \textbf{output} p-LGG}
	\label{alg:instantiation}
\end{algorithm}


Algorithm \ref{alg:instantiation} processes the lifted landmarks (nodes in p-LGG) and tries to instantiate them with a set of landmarks $Lms$,
considering the constraints in $\mathcal{V}_{csts}$. It is important to note that in this algorithm, some lifted landmarks might remain with open parameters, as the instantiation could not cover every variable in it. This is the generic case for disjunctive landmarks: if one parameter remains as a variable, it represents a disjunctive landmark.
More specifically, the algorithm behaves as follows. Line 1 initializes a map with the variable instantiation.
Lines 2-3 loop all over the landmarks of $Lms$ and search for the best equivalent predicates, i.e., those that best match in p-LGG, avoiding the constraints in $\mathcal{V}_{csts}$. 
This search requires further explanation (see below).
Line 4 initializes a queue with the lifted landmarks of \text{p-LGG}, which are processed iteratively (lines 5-11).
Line 7 tries to instantiate the parameters of the lifted landmark of line 6 by using the information on $var\_inst$.
This instantiation might be incomplete; some parameters of $lmLifted$ are instantiated, but others might remain uninstantiated. In other words, if no instantiation has been done, then $lmLifted = lmInst$. On the contrary, if some instantiation has been possible ($lmLifted \ne lmInst$), lines 9-11 update p-LGG; line 11 copies all the information of p-LGG[$lmLifted$] to the new instantiated version p-LGG[$lmInst$].

\subsubsection*{Search the best equivalent predicates}
In order to find the predicates to instantiate in line 3 of Algorithm \ref{alg:instantiation}, we first need the notion of equivalent predicates, as specified according to the following definitions.

\begin{definition}[Equivalent parameters]
\label{definition::EquivalentArgs}
Two parameters $x_1, x_2$ are equivalent, denoted as $x_1 \equiv x_2$, when:

\begin{itemize}
    \item $x_1 = x_2$, $x_1, x_2 \in  \mathcal{O}$ or $x_1, x_2 \in  \mathcal{V}$.
    \item $x_i \in \mathcal{V}$ and $x_j \in \mathcal{O}$ ($i, j \in \{1, 2\}, i \ne j$) and  $x_i$ can be instantiated with $x_j$ and $x_j \not \in \mathcal{V}_{csts}[x_i].objects$
    \item $\{x_i, x_j\} \subset \mathcal{V}$ ($i, j \in \{1, 2\}, i \ne j$), $\mathcal{V}_{csts}[x_j].objects = \mathcal{V}_{csts}[x_i].objects$  and $x_i \not \in \mathcal{V}_{csts}[x_j].variables$.
\end{itemize}
\end{definition}

\begin{definition}[Equivalent predicates]
\label{definition::Match}
    Two predicates $p_1(x_1,\dots,  x_n), p_2(y_1,\dots,  y_n)$ are equivalent, denoted as $p_1 \equiv p_2$, if $p_1 = p_2$, and $x_i \equiv y_i$, $\forall i \in \{1,\dots,  n\}$.
\end{definition}

The search for equivalent predicates is simple, but selecting the best set of equivalent predicates can be difficult. 
As an example, let us consider the predicate \texttt{p(a,b,c)} and the potential set of its equivalent predicates \{\texttt{p(a,?x0,?x1)}, \texttt{p(?x2,b,?x3)}, \texttt{p(a,?x4,c)}, \texttt{p(a,b,?x5)},\texttt{p(?x6,?x7,?x8)}\}.
We introduce a simple measure of \textit{distance} between \texttt{p(a,b,c)} and each of its equivalent predicates to decide the best set.
This distance counts the different parameters in the equivalence, i.e., possibly distinct information. For example, the distance between \texttt{p(a,b,c)} and \texttt{p(a,?x0,?x1)} is 2, but 1 between \texttt{p(a,b,c)} and \texttt{p(a,?x4,c)}. We are interested in the equivalent predicates that minimize such a distance. For this distance, we choose the top-$n$ candidates with the highest probability.
For example, let us consider that the equivalent predicates in the set are $\{\texttt{p(a,?x4,c)},\texttt{p(a,b,?x5)}\}$, with probability 0.5 and 0.25, respectively. In this case, if the number of candidates is $n=1$, only $\texttt{p(a,?x4,c)}$ is returned. If $n = 2$, both would be returned.

\section{Evaluation}




In this section, we evaluate our capabilities to extract the probabilistic ordering relationships and their (lifted) landmarks in comparison with classical methods \cite{keyder2010sound, cresswell}.  We have used two different datasets as input, namely DS1 and DS2, to learn the LGGs:

\begin{itemize}
    \item DS1. Landmark information is given by the algorithm implemented in LAMA \cite{richter2010lama}, as introduced in \cite{cresswell}. 
    This algorithm computes ordering relationships between individual+disjunctive landmarks
    and works under an SAS+ setting, which uses domain transition graphs to derive further landmarks. 
    We represent these disjunctive landmarks as lifted landmarks.
    
    \item DS2. Landmark information is given by the algorithm in \cite{keyder2010sound}, which computes a set of causal landmarks for AND/OR graphs. Combining this method with the $\Pi^m$ compilation \cite{haslum2009hm} allows for computing individual+conjunctive landmarks that consider delete information in planning problems. In our experiments, we set $m=2$, thus obtaining a set of conjunctive landmark orderings of the form $a \wedge b \rightarrow c \wedge d$, where $a, b, c, d$ are individual landmarks. We represent these landmark orderings as the single (nonconjunctive) relationships, i.e., $a \rightarrow c, a \rightarrow d, b \rightarrow c$ and $b \rightarrow d$.     
    Since our approach learns from individual landmark orderings, we decompose this conjunctive ordering to state that $a$ must happen before $c$ and $d$, and the same for $b$. However, we cannot extract an ordering between $a$ and $b$, nor $c$ and $d$. 
\end{itemize}

The evaluation process consists of three steps: 1) we extract the LGGs (with the landmark ordering relationships) provided by DS1 and DS2 over a range of planning tasks of a planning domain; 
2) we create a p-LOG, as a way to learn a lifted version of the ordering relationships of the domain; and 3) we instantiate a p-LGG for a new planning task and analyze the results \textit{w.r.t.} DS1 and DS2. 

We evaluate the quality of our results from three perspectives. First, we evaluate the running times. Second, we assess how many of our probabilistic orderings and landmarks are originally calculated by DS1/DS2.
This is an indication of the accuracy of the results, because
our probabilistic orderings are not necessarily correct, i.e., true orderings. 
Third, we assess how many of our probabilistic landmarks are true landmarks, but not calculated by DS1/DS2 (as they are not necessarily complete). 
This is an indication on how much our approach exceeds, or subsumes, the capabilities of DS1/DS2.

Note that the evaluation is limited to greedy necessary orders between landmarks, but our entire approach can deal with other orders by simply including them in the LGGs 
of the input dataset.



\subsection{Metrics}

The common metrics used in learning are inspired by those used in pattern recognition, that is, precision, recall, and their harmonic mean or F1-score. 
Comparing two landmarks or two ordering relationships is simple: if they are equal it counts as a hit, and as a failure otherwise. However, comparing a landmark and a lifted landmark, where parameters are still uninstantiated, is not so simple. Since our approach generates both landmarks and lifted landmarks in the p-LGG, we also want to calculate the \textit{similarity} between a landmark and a lifted landmark.
Consequently, we propose two new metrics to assess this similarity, the $\alpha$-precision and $\alpha$-recall.

\begin{definition}[Likelihood of landmark and lifted landmark]
    Given a lifted landmark $L'_1$ and a landmark $L_2$, being $L'_1 \equiv L_2$, 
    the likelihood of $L'_1$ and $L_2$,  denoted as $\omega(L'_1,L_2)$, is given by:
    \[\omega(L'_1,L_2) =  \begin{cases} 
      \frac{|objects(L'_1)|}{|objects(L_2)|} & |params(L_2)| > 0 \\
      1 & \text{otherwise}

   \end{cases}
    \] 

\end{definition}

\begin{definition}[Likelihood of two edges]
    Given an edge $e_1$ between two lifted landmarks ($e_1 = (L'_1, L'_2)$) and an edge $e_2$ between two landmarks ($e_2 = (L_1, L_2)$), being $L_1' \equiv L_1$ and $L_2'\equiv L_2$, 
    the likelihood of $e_1$ and $e_2$, denoted as $\omega(e_1,e_2)$, is given by:
    \[\omega(e_1,e_2) =\frac{1}{2} \cdot \left ( \omega(L'_1,L_1) + \omega(L'_2,L_2) \right )\]
\end{definition}

\begin{definition}[Likelihood of a set of landmarks  to a landmark]
Given a landmark $L$ and a set of lifted landmarks $\mathcal{L}' = \{L'_i | L'_i \equiv L\}$, the likelihood of $\mathcal{L}'$ to $L$, denoted as $\Omega(\mathcal{L}',L)$, is given by:
    \[\Omega(\mathcal{L}',L) = \frac{\displaystyle \sum_{L'_i \in \mathcal{L}'}\omega(L'_i, L)}{|\mathcal{L}'|}\]    
\end{definition}

\begin{definition}[Likelihood of a set of edges to an edge]
    Given an edge $e = (L_1,L_2)$ between two landmarks and 
    a set of edges between two lifted landmarks $\mathcal{E} = \{e_i = (L'_1, L'_2)| L'_1 \equiv L_1 \land L'_2 \equiv L_2\}$, the likelihood of $\mathcal{E}$ to $e$, denoted as $\Omega(\mathcal{E}, e)$, is given by:
    \[\Omega(\mathcal{E}, e) = \frac{\displaystyle \sum_{e_i \in \mathcal{E}} \omega(e_i, e )}{|\mathcal{E}|}\]
\end{definition}

\begin{definition}[Likelihood of p-LGG to LGG]
\label{def:likelihood_p-LGG}
    Let us consider an LGG, given by $G = (V, E)$, and a p-LGG, given by $G_p = (V_p, E_p)$. Let $V_{diff}= V \setminus V_p$ and $E_{diff} =  E \setminus E_p$, be the vertex (landmark) and edge (ordering) difference, respectively. We define the vertex likelihood of p-LGG to LGG as
    \[\alpha_V = \frac{\displaystyle \sum_{L_i \in V_{diff}} \Omega(lifted(V_p \setminus V), L_i)}{|V_{diff}|}\]
    and the edge likelihood of p-LGG to LGG as
    \[\alpha_E = \frac{\displaystyle \sum_{e_i \in E_{diff}} \Omega(lifted(E_p \setminus E), e_i)}{|E_{diff}|}\]
    where $lifted(V) = \{L' \in V | L' \text{ is lifted}\}$ and $lifted(E) = \{(L'_1, L'_2) \in E | L'_1 \text{ is lifted} \lor L'_2 \text{ is lifted}\}$ .
    The likelihood of p-LGG to LGG is denoted as the tuple $\langle \alpha_V, \alpha_E \rangle$.
\end{definition}

We present a clarifying example for Definition \ref{def:likelihood_p-LGG}. Let us consider the LGG and p-LGG such that $V = \{$\texttt{on(b,a)}, \texttt{on(c,d)}, \texttt{ontable(a)}, \texttt{ontable(d)}$\}$ and $ V_p = \{$\texttt{on(c,d)}, \texttt{ontable(d)}, \texttt{on(b,?x0)}, \texttt{ontable(?x0)}$\}$, where \texttt{?x0} represents a variable that can be instantiated with value \texttt{a}. Therefore, from the definition, $ V_{diff} = \{$\texttt{on(b,a)}, \texttt{ontable(a)}$\}$. 
The likelihood  evaluates $ \Omega($\texttt{on(b,a)}, \{\texttt{on(b,?x0)}, \texttt{ontable(?x0)}\}$)$ which 
simplifies to
$\Omega($\texttt{on(b,a)}, \{\texttt{on(b,?x0)}$\})$, because we only have the equivalence \texttt{on(b,?x0)} $\equiv$ \texttt{on(b,a)}. To compute $\Omega$, we iterate for every item of the set computing $\omega$, obtaining:
\[
\omega(\texttt{on(b,a)}, \texttt{on(b,?x0)}) = \dfrac{1}{2}
\]
leading to
\[
\Omega(\texttt{on(B,A)}, \{\texttt{on(b,?x0)}\}) = \dfrac{\frac{1}{2}}{1} = \dfrac{1}{2}
\]
 Following the same process with \texttt{ontable(a)}, we obtain $\Omega(\texttt{ontable(a)}, \{\texttt{ontable(?x0)}\}) = \frac{1}{2}$. Consequently:
\[
\alpha_V = \dfrac{\frac{1}{2} + \frac{1}{2}}{2} = \dfrac{1}{2}
\]
Therefore, the $\alpha_V$ of LGG and p-LGG is 0.5. 
The procedure to compute the likelihood of edges is analogous. 

We propose two new metrics based on the classical precision and recall, where $\alpha$ is either $\alpha_V$ or $\alpha_E$:
\[\alpha\text{-Precision} = \text{Precision} + \alpha(1-\text{Precision})\] 
\[\alpha\text{-Recall} = \text{Recall} + \alpha(1-\text{Recall})\] 
These new metrics assess not only the classical precision and recall, which compare the grounded landmarks,
but also the similarity between landmarks and lifted landmarks.
Note that if the p-LGG does not contain lifted landmarks, the $\alpha$-metrics are analogous to the classical ones because $\alpha_V = \alpha_E = 0$.

\subsection{Setup}


The algorithms of our approach have been implemented in \texttt{Python 3.12.3}. All the experiments are run on a 32Gb RAM, i9-13900H computer. 
We have used the following 11 IPC domains: Barman, BlocksWorld, Depots, Driverlog, Elevator, Floortile, Freecell, Grid, Rovers, Satellite and ZenoTravel.\footnote{The domains are available in \newline\texttt{https://www.icaps-conference.org/competitions}}
These domains were selected based on the density of the greedy necessary landmarks orderings they contain.
While BlocksWorld, Depots and Barman are well-known for containing many orderings, Driverlog and ZenoTravel typically contain few orderings.
The remaining domains offer a broader spectrum, representing different degrees of ordering density. In every domain, we use 14 different planning tasks; randomly, 4 are used to compute the p-LOG (training) and 10 to instantiate the p-LGG and get the results (testing). We run 10 different experiments per domain, 5 from DS1 and 5 from DS2.
Therefore, the results shown in this section are the mean for each of these 5 experiments. DS2 does not get landmark information for Driverlog, which is denoted as ``-''.

\begin{table*}[h]
\centering
\caption{Precision (P), Recall (R) and F1-score (F1) for the grounded results, and their $\alpha$-versions for the lifted results, when using DS1 and DS2 as an input. Orderings are on the left and landmarks are on the right.}
\resizebox{0.8\textwidth}{!}{
\begin{tabular}{ll|llll|llll|}
\cline{3-10}
                                                  &    & \multicolumn{4}{c|}{Orderings}                                                                   & \multicolumn{4}{c|}{Landmarks}                                                                   \\ \cline{3-10} 
                                                  &    & \multicolumn{2}{c|}{Grounded}                                 & \multicolumn{2}{c|}{Lifted}              & \multicolumn{2}{c|}{Grounded}                                 & \multicolumn{2}{c|}{Lifted}              \\ \cline{3-10} 
                                                  &    & \multicolumn{1}{l|}{p-LGG$_{DS1}$} & \multicolumn{1}{l|}{p-LGG$_{DS2}$} & \multicolumn{1}{l|}{p-LGG$_{DS1}$} & p-LGG$_{DS2}$ & \multicolumn{1}{l|}{p-LGG$_{DS1}$} & \multicolumn{1}{l|}{p-LGG$_{DS2}$} & \multicolumn{1}{l|}{p-LGG$_{DS1}$} & p-LGG$_{DS2}$ \\ \hline
\multicolumn{1}{|l|}{\multirow{3}{*}{Barman}}     & P  & \multicolumn{1}{l|}{0.93}     & \multicolumn{1}{l|}{0.95}     & \multicolumn{1}{l|}{0.96}     & 0.97     & \multicolumn{1}{l|}{0.45}     & \multicolumn{1}{l|}{0.48}     & \multicolumn{1}{l|}{0.63}     & 0.70     \\ \cline{2-10} 
\multicolumn{1}{|l|}{}                            & R  & \multicolumn{1}{l|}{0.35}     & \multicolumn{1}{l|}{0.28}     & \multicolumn{1}{l|}{0.62}     & 0.65     & \multicolumn{1}{l|}{0.74}     & \multicolumn{1}{l|}{0.73}     & \multicolumn{1}{l|}{0.82}     & 0.85     \\ \cline{2-10} 
\multicolumn{1}{|l|}{}                            & F1 & \multicolumn{1}{l|}{0.51}     & \multicolumn{1}{l|}{0.44}     & \multicolumn{1}{l|}{0.75}     & 0.78     & \multicolumn{1}{l|}{0.56}     & \multicolumn{1}{l|}{0.59}     & \multicolumn{1}{l|}{0.71}     & 0.77     \\ \hline
\multicolumn{1}{|l|}{\multirow{3}{*}{BlocksWorld}}     & P  & \multicolumn{1}{l|}{0.72}     & \multicolumn{1}{l|}{0.68}     & \multicolumn{1}{l|}{0.87}     & 0.82     & \multicolumn{1}{l|}{0.80}     & \multicolumn{1}{l|}{0.84}     & \multicolumn{1}{l|}{1}        & 1        \\ \cline{2-10} 
\multicolumn{1}{|l|}{}                            & R  & \multicolumn{1}{l|}{0.91}     & \multicolumn{1}{l|}{0.75}     & \multicolumn{1}{l|}{0.96}     & 0.86     & \multicolumn{1}{l|}{1}        & \multicolumn{1}{l|}{1}        & \multicolumn{1}{l|}{1}        & 1        \\ \cline{2-10} 
\multicolumn{1}{|l|}{}                            & F1 & \multicolumn{1}{l|}{0.80}     & \multicolumn{1}{l|}{0.72}     & \multicolumn{1}{l|}{0.91}     & 0.84     & \multicolumn{1}{l|}{0.89}     & \multicolumn{1}{l|}{0.91}     & \multicolumn{1}{l|}{1}        & 1        \\ \hline
\multicolumn{1}{|l|}{\multirow{3}{*}{Depots}}     & P  & \multicolumn{1}{l|}{0.65}     & \multicolumn{1}{l|}{0.15}     & \multicolumn{1}{l|}{0.83}     & 0.60     & \multicolumn{1}{l|}{0.63}     & \multicolumn{1}{l|}{0.60}     & \multicolumn{1}{l|}{0.67}     & 0.63     \\ \cline{2-10} 
\multicolumn{1}{|l|}{}                            & R  & \multicolumn{1}{l|}{0.23}     & \multicolumn{1}{l|}{0.09}     & \multicolumn{1}{l|}{0.62}     & 0.56     & \multicolumn{1}{l|}{0.83}     & \multicolumn{1}{l|}{0.47}     & \multicolumn{1}{l|}{0.85}     & 0.51     \\ \cline{2-10} 
\multicolumn{1}{|l|}{}                            & F1 & \multicolumn{1}{l|}{0.35}     & \multicolumn{1}{l|}{0.11}     & \multicolumn{1}{l|}{0.71}     & 0.58     & \multicolumn{1}{l|}{0.71}     & \multicolumn{1}{l|}{0.53}     & \multicolumn{1}{l|}{0.75}     & 0.56     \\ \hline
\multicolumn{1}{|l|}{\multirow{3}{*}{Driverlog}}  & P  & \multicolumn{1}{l|}{0}        & \multicolumn{1}{l|}{-}        & \multicolumn{1}{l|}{0.54}     & -        & \multicolumn{1}{l|}{0.26}     & \multicolumn{1}{l|}{-}        & \multicolumn{1}{l|}{0.41}     & -        \\ \cline{2-10} 
\multicolumn{1}{|l|}{}                            & R  & \multicolumn{1}{l|}{1}        & \multicolumn{1}{l|}{-}        & \multicolumn{1}{l|}{1}        & -        & \multicolumn{1}{l|}{1}        & \multicolumn{1}{l|}{-}        & \multicolumn{1}{l|}{1}        & -        \\ \cline{2-10} 
\multicolumn{1}{|l|}{}                            & F1 & \multicolumn{1}{l|}{0}        & \multicolumn{1}{l|}{-}        & \multicolumn{1}{l|}{0.70}     & -        & \multicolumn{1}{l|}{0.41}     & \multicolumn{1}{l|}{-}        & \multicolumn{1}{l|}{0.58}     & -        \\ \hline
\multicolumn{1}{|l|}{\multirow{3}{*}{Elevator}}   & P  & \multicolumn{1}{l|}{0.92}     & \multicolumn{1}{l|}{0.93}     & \multicolumn{1}{l|}{0.96}     & 0.97     & \multicolumn{1}{l|}{0.10}     & \multicolumn{1}{l|}{0.08}     & \multicolumn{1}{l|}{0.10}     & 0.09     \\ \cline{2-10} 
\multicolumn{1}{|l|}{}                            & R  & \multicolumn{1}{l|}{0.34}     & \multicolumn{1}{l|}{0.34}     & \multicolumn{1}{l|}{0.66}     & 0.67     & \multicolumn{1}{l|}{0.63}     & \multicolumn{1}{l|}{0.63}     & \multicolumn{1}{l|}{0.63}     & 0.63     \\ \cline{2-10} 
\multicolumn{1}{|l|}{}                            & F1 & \multicolumn{1}{l|}{0.50}     & \multicolumn{1}{l|}{0.50}     & \multicolumn{1}{l|}{0.79}     & 0.79     & \multicolumn{1}{l|}{0.18}     & \multicolumn{1}{l|}{0.15}     & \multicolumn{1}{l|}{0.18}     & 0.15     \\ \hline
\multicolumn{1}{|l|}{\multirow{3}{*}{Floortile}}  & P  & \multicolumn{1}{l|}{0.69}     & \multicolumn{1}{l|}{0.80}     & \multicolumn{1}{l|}{0.86}     & 1        & \multicolumn{1}{l|}{0.29}     & \multicolumn{1}{l|}{0.31}     & \multicolumn{1}{l|}{0.32}     & 1        \\ \cline{2-10} 
\multicolumn{1}{|l|}{}                            & R  & \multicolumn{1}{l|}{1}        & \multicolumn{1}{l|}{1}        & \multicolumn{1}{l|}{1}        & 1        & \multicolumn{1}{l|}{1}        & \multicolumn{1}{l|}{1}        & \multicolumn{1}{l|}{1}        & 1        \\ \cline{2-10} 
\multicolumn{1}{|l|}{}                            & F1 & \multicolumn{1}{l|}{0.82}     & \multicolumn{1}{l|}{0.89}     & \multicolumn{1}{l|}{0.92}     & 1        & \multicolumn{1}{l|}{0.46}     & \multicolumn{1}{l|}{0.47}     & \multicolumn{1}{l|}{0.48}     & 1        \\ \hline
\multicolumn{1}{|l|}{\multirow{3}{*}{Freecell}}   & P  & \multicolumn{1}{l|}{0.76}     & \multicolumn{1}{l|}{0.53}     & \multicolumn{1}{l|}{0.83}     & 0.69     & \multicolumn{1}{l|}{0.28}     & \multicolumn{1}{l|}{0.35}     & \multicolumn{1}{l|}{0.29}     & 0.36     \\ \cline{2-10} 
\multicolumn{1}{|l|}{}                            & R  & \multicolumn{1}{l|}{0.26}     & \multicolumn{1}{l|}{0.32}     & \multicolumn{1}{l|}{0.48}     & 0.55     & \multicolumn{1}{l|}{0.70}     & \multicolumn{1}{l|}{0.73}     & \multicolumn{1}{l|}{0.70}     & 0.73     \\ \cline{2-10} 
\multicolumn{1}{|l|}{}                            & F1 & \multicolumn{1}{l|}{0.39}     & \multicolumn{1}{l|}{0.39}     & \multicolumn{1}{l|}{0.61}     & 0.61     & \multicolumn{1}{l|}{0.41}     & \multicolumn{1}{l|}{0.48}     & \multicolumn{1}{l|}{0.41}     & 0.48     \\ \hline
\multicolumn{1}{|l|}{\multirow{3}{*}{Grid}}       & P  & \multicolumn{1}{l|}{0.40}     & \multicolumn{1}{l|}{0.11}     & \multicolumn{1}{l|}{0.60}     & 0.54     & \multicolumn{1}{l|}{0.06}     & \multicolumn{1}{l|}{0.05}     & \multicolumn{1}{l|}{0.13}     & 1        \\ \cline{2-10} 
\multicolumn{1}{|l|}{}                            & R  & \multicolumn{1}{l|}{0.75}     & \multicolumn{1}{l|}{0.76}     & \multicolumn{1}{l|}{0.83}     & 0.88     & \multicolumn{1}{l|}{0.92}     & \multicolumn{1}{l|}{1}        & \multicolumn{1}{l|}{0.92}     & 1        \\ \cline{2-10} 
\multicolumn{1}{|l|}{}                            & F1 & \multicolumn{1}{l|}{0.52}     & \multicolumn{1}{l|}{0.19}     & \multicolumn{1}{l|}{0.69}     & 0.67     & \multicolumn{1}{l|}{0.11}     & \multicolumn{1}{l|}{0.11}     & \multicolumn{1}{l|}{0.24}     & 1        \\ \hline
\multicolumn{1}{|l|}{\multirow{3}{*}{Rovers}}     & P  & \multicolumn{1}{l|}{0.10}     & \multicolumn{1}{l|}{0.66}     & \multicolumn{1}{l|}{0.67}     & 0.83     & \multicolumn{1}{l|}{0.07}     & \multicolumn{1}{l|}{0.11}     & \multicolumn{1}{l|}{0.34}     & 0.22     \\ \cline{2-10} 
\multicolumn{1}{|l|}{}                            & R  & \multicolumn{1}{l|}{0.89}     & \multicolumn{1}{l|}{0.06}     & \multicolumn{1}{l|}{0.83}     & 0.29     & \multicolumn{1}{l|}{0.89}     & \multicolumn{1}{l|}{0.70}     & \multicolumn{1}{l|}{0.93}     & 0.73     \\ \cline{2-10} 
\multicolumn{1}{|l|}{}                            & F1 & \multicolumn{1}{l|}{0.14}     & \multicolumn{1}{l|}{0.14}     & \multicolumn{1}{l|}{0.74}     & 0.43     & \multicolumn{1}{l|}{0.14}     & \multicolumn{1}{l|}{0.19}     & \multicolumn{1}{l|}{0.49}     & 0.34     \\ \hline
\multicolumn{1}{|l|}{\multirow{3}{*}{Satellite}}  & P  & \multicolumn{1}{l|}{0.28}     & \multicolumn{1}{l|}{0.89}     & \multicolumn{1}{l|}{0.69}     & 0.97     & \multicolumn{1}{l|}{0.25}     & \multicolumn{1}{l|}{0.44}     & \multicolumn{1}{l|}{0.50}     & 0.52     \\ \cline{2-10} 
\multicolumn{1}{|l|}{}                            & R  & \multicolumn{1}{l|}{0.67}     & \multicolumn{1}{l|}{0.11}     & \multicolumn{1}{l|}{0.84}     & 0.31     & \multicolumn{1}{l|}{0.90}     & \multicolumn{1}{l|}{0.70}     & \multicolumn{1}{l|}{0.93}     & 0.73     \\ \cline{2-10} 
\multicolumn{1}{|l|}{}                            & F1 & \multicolumn{1}{l|}{0.40}     & \multicolumn{1}{l|}{0.19}     & \multicolumn{1}{l|}{0.76}     & 0.47     & \multicolumn{1}{l|}{0.39}     & \multicolumn{1}{l|}{0.54}     & \multicolumn{1}{l|}{0.65}     & 0.61     \\ \hline
\multicolumn{1}{|l|}{\multirow{3}{*}{ZenoTravel}} & P  & \multicolumn{1}{l|}{0.05}     & \multicolumn{1}{l|}{1}        & \multicolumn{1}{l|}{0.54}     & 1        & \multicolumn{1}{l|}{0.52}     & \multicolumn{1}{l|}{0.46}     & \multicolumn{1}{l|}{0.68}     & 0.46     \\ \cline{2-10} 
\multicolumn{1}{|l|}{}                            & R  & \multicolumn{1}{l|}{0.95}     & \multicolumn{1}{l|}{0}        & \multicolumn{1}{l|}{0.97}     & 0        & \multicolumn{1}{l|}{1}        & \multicolumn{1}{l|}{0.75}     & \multicolumn{1}{l|}{1}        & 0.75     \\ \cline{2-10} 
\multicolumn{1}{|l|}{}                            & F1 & \multicolumn{1}{l|}{0.10}     & \multicolumn{1}{l|}{0}        & \multicolumn{1}{l|}{0.69}     & 0        & \multicolumn{1}{l|}{0.69}     & \multicolumn{1}{l|}{0.57}       & \multicolumn{1}{l|}{0.81}     & 0.57       \\ \hline
\end{tabular}}
\label{tab:comparisonBaseline}
\end{table*}

\begin{table}[t]
\centering
\caption{Precision (P), Recall (R) and F1-score (F1) for the grounded landmarks when comparing the landmarks extracted by p-LGG$_{DS1}$, DS1, p-LGG$_{DS2}$ and DS2 \textit{w.r.t.} the baseline.}
\resizebox{0.35\textwidth}{!}{
\begin{tabular}{ll|llll|}
\cline{3-6}
                                                  &    & \multicolumn{4}{c|}{Landmarks}                                                                                   \\ \cline{3-6} 
                                                  &    & \multicolumn{1}{l|}{p-LGG$_{DS1}$} & \multicolumn{1}{l|}{DS1} & \multicolumn{1}{l|}{p-LGG$_{DS2}$} & DS2   \\ \hline
\multicolumn{1}{|l|}{\multirow{3}{*}{Barman}}     & P  & \multicolumn{1}{l|}{0.97}              & \multicolumn{1}{l|}{1}    & \multicolumn{1}{l|}{0.98}            & 1    \\ \cline{2-6} 
\multicolumn{1}{|l|}{}                            & R  & \multicolumn{1}{l|}{0.85}              & \multicolumn{1}{l|}{0.53} & \multicolumn{1}{l|}{0.85}            & 0.56 \\ \cline{2-6} 
\multicolumn{1}{|l|}{}                            & F1 & \multicolumn{1}{l|}{0.90}              & \multicolumn{1}{l|}{0.69} & \multicolumn{1}{l|}{0.91}            & 0.73 \\ \hline
\multicolumn{1}{|l|}{\multirow{3}{*}{BlocksWorld}}     & P  & \multicolumn{1}{l|}{0.85}              & \multicolumn{1}{l|}{1}    & \multicolumn{1}{l|}{0.85}            & 1    \\ \cline{2-6} 
\multicolumn{1}{|l|}{}                            & R  & \multicolumn{1}{l|}{1}                 & \multicolumn{1}{l|}{0.93} & \multicolumn{1}{l|}{1}               & 1    \\ \cline{2-6} 
\multicolumn{1}{|l|}{}                            & F1 & \multicolumn{1}{l|}{0.92}              & \multicolumn{1}{l|}{0.97} & \multicolumn{1}{l|}{0.92}            & 1    \\ \hline
\multicolumn{1}{|l|}{\multirow{3}{*}{Depots}}     & P  & \multicolumn{1}{l|}{0.96}              & \multicolumn{1}{l|}{1}    & \multicolumn{1}{l|}{0.92}            & 1    \\ \cline{2-6} 
\multicolumn{1}{|l|}{}                            & R  & \multicolumn{1}{l|}{0.80}              & \multicolumn{1}{l|}{0.63} & \multicolumn{1}{l|}{0.81}            & 0.68 \\ \cline{2-6} 
\multicolumn{1}{|l|}{}                            & F1 & \multicolumn{1}{l|}{0.87}              & \multicolumn{1}{l|}{0.77} & \multicolumn{1}{l|}{0.86}            & 0.81 \\ \hline
\multicolumn{1}{|l|}{\multirow{3}{*}{Driverlog}}  & P  & \multicolumn{1}{l|}{0.95}              & \multicolumn{1}{l|}{1}    & \multicolumn{1}{l|}{-}               & -    \\ \cline{2-6} 
\multicolumn{1}{|l|}{}                            & R  & \multicolumn{1}{l|}{1}                 & \multicolumn{1}{l|}{0.29} & \multicolumn{1}{l|}{-}               & -    \\ \cline{2-6} 
\multicolumn{1}{|l|}{}                            & F1 & \multicolumn{1}{l|}{0.97}              & \multicolumn{1}{l|}{0.45} & \multicolumn{1}{l|}{-}               & -    \\ \hline
\multicolumn{1}{|l|}{\multirow{3}{*}{Elevator}}   & P  & \multicolumn{1}{l|}{1}                 & \multicolumn{1}{l|}{1}    & \multicolumn{1}{l|}{1}               & 1    \\ \cline{2-6} 
\multicolumn{1}{|l|}{}                            & R  & \multicolumn{1}{l|}{0.94}              & \multicolumn{1}{l|}{0.14} & \multicolumn{1}{l|}{0.94}            & 0.16 \\ \cline{2-6} 
\multicolumn{1}{|l|}{}                            & F1 & \multicolumn{1}{l|}{0.97}              & \multicolumn{1}{l|}{0.25} & \multicolumn{1}{l|}{0.97}            & 0.26 \\ \hline
\multicolumn{1}{|l|}{\multirow{3}{*}{Floortile}}  & P  & \multicolumn{1}{l|}{0.95}              & \multicolumn{1}{l|}{1}    & \multicolumn{1}{l|}{0.94}            & 1    \\ \cline{2-6} 
\multicolumn{1}{|l|}{}                            & R  & \multicolumn{1}{l|}{0.99}              & \multicolumn{1}{l|}{0.33} & \multicolumn{1}{l|}{0.99}            & 0.33 \\ \cline{2-6} 
\multicolumn{1}{|l|}{}                            & F1 & \multicolumn{1}{l|}{0.97}              & \multicolumn{1}{l|}{0.49} & \multicolumn{1}{l|}{0.97}            & 0.49 \\ \hline
\multicolumn{1}{|l|}{\multirow{3}{*}{Freecell}}   & P  & \multicolumn{1}{l|}{1}                 & \multicolumn{1}{l|}{1}    & \multicolumn{1}{l|}{1}               & 1    \\ \cline{2-6} 
\multicolumn{1}{|l|}{}                            & R  & \multicolumn{1}{l|}{0.88}              & \multicolumn{1}{l|}{0.36} & \multicolumn{1}{l|}{0.89}            & 0.43 \\ \cline{2-6} 
\multicolumn{1}{|l|}{}                            & F1 & \multicolumn{1}{l|}{0.94}              & \multicolumn{1}{l|}{0.53} & \multicolumn{1}{l|}{0.94}            & 0.60 \\ \hline
\multicolumn{1}{|l|}{\multirow{3}{*}{Grid}}       & P  & \multicolumn{1}{l|}{0.79}              & \multicolumn{1}{l|}{1}    & \multicolumn{1}{l|}{0.80}            & 1    \\ \cline{2-6} 
\multicolumn{1}{|l|}{}                            & R  & \multicolumn{1}{l|}{0.99}              & \multicolumn{1}{l|}{0.05} & \multicolumn{1}{l|}{0.99}            & 0.07 \\ \cline{2-6} 
\multicolumn{1}{|l|}{}                            & F1 & \multicolumn{1}{l|}{0.89}              & \multicolumn{1}{l|}{0.09} & \multicolumn{1}{l|}{0.89}            & 0.13 \\ \hline
\multicolumn{1}{|l|}{\multirow{3}{*}{Rovers}}     & P  & \multicolumn{1}{l|}{0.94}              & \multicolumn{1}{l|}{1}    & \multicolumn{1}{l|}{0.93}            & 1    \\ \cline{2-6} 
\multicolumn{1}{|l|}{}                            & R  & \multicolumn{1}{l|}{0.98}              & \multicolumn{1}{l|}{0.10} & \multicolumn{1}{l|}{0.99}            & 0.15 \\ \cline{2-6} 
\multicolumn{1}{|l|}{}                            & F1 & \multicolumn{1}{l|}{0.96}              & \multicolumn{1}{l|}{0.17} & \multicolumn{1}{l|}{0.96}            & 0.23 \\ \hline
\multicolumn{1}{|l|}{\multirow{3}{*}{Satellite}}  & P  & \multicolumn{1}{l|}{0.96}              & \multicolumn{1}{l|}{1}    & \multicolumn{1}{l|}{0.97}            & 1    \\ \cline{2-6} 
\multicolumn{1}{|l|}{}                            & R  & \multicolumn{1}{l|}{0.95}              & \multicolumn{1}{l|}{0.28} & \multicolumn{1}{l|}{0.88}            & 0.44 \\ \cline{2-6} 
\multicolumn{1}{|l|}{}                            & F1 & \multicolumn{1}{l|}{0.95}              & \multicolumn{1}{l|}{0.44} & \multicolumn{1}{l|}{0.92}            & 0.61 \\ \hline
\multicolumn{1}{|l|}{\multirow{3}{*}{ZenoTravel}} & P  & \multicolumn{1}{l|}{0.99}              & \multicolumn{1}{l|}{1}    & \multicolumn{1}{l|}{1}               & 1    \\ \cline{2-6} 
\multicolumn{1}{|l|}{}                            & R  & \multicolumn{1}{l|}{1}                 & \multicolumn{1}{l|}{0.49} & \multicolumn{1}{l|}{0.86}            & 0.87 \\ \cline{2-6} 
\multicolumn{1}{|l|}{}                            & F1 & \multicolumn{1}{l|}{0.99}              & \multicolumn{1}{l|}{0.65} & \multicolumn{1}{l|}{0.93}            & 0.93 \\ \hline
\end{tabular}}
\label{tab:landmarkRetrieval}
\end{table}

\subsection{Results}

Firstly, we evaluate the execution time of our algorithms. The times to learn the p-LOG take from 0.05s to 0.30s and 0.04s to 0.33s when using DS1 and DS2 as input, respectively. The times to instantiate the p-LGG take from 0.37s to 3.37s and 0.52s to 3.33s, respectively. On the other hand, DS1 takes from 0.1s up to 1.5s to compute the LGG, whereas DS2 takes from 1s up to several hours for some tasks, depending on the domain and task. Our approach takes up to 2s more than DS1 to instantiate the landmark information, but it is significantly faster than DS2, which takes up to several days.

Secondly, we evaluate how similar the results generated from our p-LGG (namely p-LGG$_{DS1}$ and p-LGG$_{DS2}$) are to the original LGGs calculated by DS1 and DS2.
We perform here a \textit{grounded} landmark-to-landmark comparison and a \textit{lifted} landmark-to-disjunctive landmark comparison. 
Table \ref{tab:comparisonBaseline} shows the average results for these comparisons.
For the grounded columns, the table represents the precision, recall and F1-score, whereas for the lifted columns, it represents 
the $\alpha$-precision, $\alpha$-recall and the $\alpha$-F1-score. 
Higher values in the table represent better results; a value of 1 means that we obtain the
perfect results. 
The results are heavily influenced by the domain and their known difficulty in providing landmarks \cite{marzal2011full}. Barman, BlocksWorld and Floortile, with more landmarks,
show the best results. Other domains, such as Driverlog, Grid, Rovers and ZenoTravel, with fewer landmarks, show more irregular results. 
Intuitively, the results for the orderings should be worse than for the landmarks; after all, the orderings always require the two landmarks to be correct to be counted as a hit.
This happens for recall, which is oriented toward completeness. However, in terms of precision, which is oriented to positive predictiveness, this is not always the case, and the results for the orderings may be better than for the landmarks. This is due to the difference between the number of orderings \textit{vs.} the number of landmarks: there are fewer orderings than landmarks. Therefore, the proportion of positive orderings may be higher than the proportion of positive landmarks.
In general, the results are better for the lifted columns than for the grounded columns. 
There is a reason for this: since we instantiate the learned information from the domain in a probabilistic way (p-LOG), we include orderings with a probability lower than 1, resulting in an instantiated p-LGG that tends to include more landmarks and orderings than it should. 
In some domains, there are clear differences between p-LGG$_{DS1}$ and p-LGG$_{DS2}$. 
Particularly in ZenoTravel, the LGGs in DS2 hardly extract greedy necessary orderings, which negatively affects the p-LOG learning and the p-LGG instantiation. In other domains, such as Depots, Grid and Satellite, we have not yet discovered the actual reason behind these differences, but our intuition is that some of the conjunctive orderings worsen the p-LOG learning process.

Thirdly, we evaluate how much extra information, beyond DS1/DS2, our p-LGG generates. 
DS1 does not extract all the possible landmarks of a planning task \cite{hoffmann2004ordered,richter2010lama}. DS2 may extract all possible landmarks, if $m$ is big enough and without time limitations \cite{keyder2010sound}, which could be computationally untractable.
In this experiment, we define a baseline that computes all the true landmarks. It is calculated by using a brute force algorithm that exhaustively builds a relaxed planning graph to check whether a predicate is a landmark or not. This process extracts the landmarks but not their ordering relationship.
On the other hand, DS2 does not calculate disjunctive/lifted landmarks. Consequently, we can only perform a grounded landmark-to-landmark comparison here.
Table \ref{tab:landmarkRetrieval} shows the average results of p-LGG$_{DS1}$, DS1, p-LGG$_{DS2}$ and DS2 \textit{w.r.t.} the baseline.
Higher values in the table represent better results; a value of 1 means that the method obtains the perfect results.
Higher values in p-LGG$_{DS1}$ and p-LGG$_{DS2}$ \textit{vs.} DS1 and DS2, respectively, mean that our approach extracts true landmarks that DS1/DS2 fail to extract. 
Obviously, the precision of DS1/DS2 is always 1, as all their landmarks are true landmarks. Our approach does not guarantee true landmarks, but the precision is over 0.9 in 9 out of 11 domains.
From the recall perspective, 
our approach is clearly better as it always extracts more landmarks than in the original DS1/DS2, except in ZenoTravel.
This confirms that
leveraging domain-learned information enables to extract more landmarks than classical methods. 



\section{Conclusions through related work}


Works in the literature primarily focus on the definition of landmark-counting heuristics for lifted planning. In \cite{wichlacz2022landmark}, a new family of landmark heuristics derived from the usual $h^{LC}$ is introduced. Similarly, \cite{wichlacz2023landmark} presents a heuristic for optimal lifted planning based on the LM cut heuristic.
In \cite{schubertreasonable}, a way to generalize the concept of reasonable landmark orderings 
to improve the search is proposed. 
Regarding new landmark-like methods, \cite{kim2024relevance} explores a new concept of \textit{relevance score} to compute the relevance of a given predicate: although one predicate might not be a true landmark, it can be useful in the heuristic search if its relevance score is high. 
This is aligned with our intuition on the landmarks present in a probabilistic ordering relationship; they might not be true, but they are still relevant.
In a similar direction, \cite{molina2020learn} provides a different approximation for motion planning in the form of \textit{critical regions} on the topology of the domain, which can help achieve the goals faster.

To 
our knowledge, there are no significant advances to 
explore new ways to 
generate lifted landmarks and their orderings beyond the classical methods. This work 
proposes a novel method for this by capturing probabilistic landmark orderings (from an input of planning tasks) that are lifted and, thereby, applicable across the entire domain. 
The main contributions of this work are: 1) how to generalize domain-wide ordering relationships between landmarks associated with a probability for domain-landmark learning; and 2)
how to particularize the information of the domain to a new planning task, for problem-landmark generation. 
We have shown that learning lifted landmarks and their relationships from just a few instances (we only use four) allows us to capture valuable probabilistic information on the domain. With that information and a new planning task, we instantiate the landmarks orderings.
The strength of this work relies on the potential of the learned lifted landmarks, which allow us to instantiate landmarks that are missing under other methods. Our experiments show good results for this, particularly in terms of the recall. This reinforces the interest in landmark generalization and opens new trends for exploring lifted landmarks and their relationships to specific instances.
The weakness of the work relies on the probabilistic learning behavior, which might instantiate incorrect (i.e., untrue) orderings and landmarks in some domains, particularly in those with fewer landmarks and with many open parameters. This is shown in some domains by low precision values.

As future work, we want to analyze how to safely combine greedy necessary orders with other orders, such as reasonable ones, 
in a probabilistic way. 
In order to improve the precision, we want to study alternatives (similar to relevance score or critical regions) to anticipate whether a probabilistic landmark ordering is more valuable for one planning task or another. Furthermore, we are currently investigating methods to extract lifted conjunctive and mutex relationships between orderings based on the input dataset. Advancements in this direction can potentially enhance the quality of results. Finally, since landmarks serve as a means to guide search processes, we are also exploring the use of instantiated p-LGGs in a heuristic-like manner, with the aim of directing the search more effectively.
This is crucial to analyze when a planner can benefit from the generated information and if an incorrect probabilistic landmark has a negative impact on the search.
The main difficulty here is that this analysis depends entirely on the planner and how it uses the landmark information.


\bibliographystyle{IEEEtran}
\bibliography{aaai25}
\end{document}